\documentclass[twocolumn]{article}
\usepackage[utf8]{inputenc}
\usepackage{multirow}
\usepackage{graphicx}
\usepackage[colorlinks=true,linkcolor=blue,citecolor=blue]{hyperref}
\usepackage{float}
\usepackage[a4paper, margin=2cm]{geometry}
\usepackage{amsmath}
\usepackage{amsfonts}
\usepackage{amssymb}
\usepackage{verbatim} 
\usepackage{float} 
\usepackage{hyperref} 
\usepackage{authblk}

\author[1]{Celina Hanouti}
\author[2]{Hervé Le Borgne\footnote{corresponding author:  \href{mailto: herve.le-borgne@cea.fr}{herve.le-borgne@cea.fr}}}
\affil[1]{Wefox}
\affil[2]{Université Paris-Saclay, CEA, List, F-91120, Palaiseau, France.}
\title{Learning Semantic Ambiguities for Zero-Shot Learning}
\date{}
\begin{document}
\maketitle

\begin{abstract}
Zero-shot learning (ZSL) aims at recognizing classes for which no visual sample is available at training time. 
To address this issue, one can rely on a semantic description of each class. A typical ZSL model learns a mapping between the visual samples of seen classes and the corresponding semantic descriptions, in order to do the same on unseen classes at test time.
State of the art approaches rely on generative models that synthesize visual features from the prototype of a class, such that a classifier can then be learned in a supervised manner. However, these approaches are usually biased towards seen classes whose visual instances are the only one that can be matched to a given class prototype.
We propose a regularization method that can be applied to any conditional generative-based ZSL method, by leveraging only the semantic class prototypes.
It learns to synthesize discriminative features for possible semantic description that are not available at training time, that is the unseen ones. 
The approach is evaluated for ZSL and GZSL on four datasets commonly used in the literature, either in inductive and transductive settings, with results on-par or above state of the art approaches.
\end{abstract}

\section{Introduction}
Being able to classify, detect or segment objects into images with as less annotated data as possible is one of the most important problem addressed to implement practical application in computer vision. 
A radical framework is proposed by the zero-shot learning (ZSL), in which not a single visual example is used during learning, but where it is possible to rely on external data from another modality. Typically, the latter are semantic attributes or textual descriptions that can be represented by vectors. Hence, the task consists in learning a mapping between the image space and the semantic space using images from seen classes, available at training time only. In the original form of ZSL, the images of the test set belong to unseen classes, for which no sample is available at training time. A more realistic ``generalized'' setting (GZSL)~\cite{xian2017} proposes nevertheless to recognize both seen and unseen classes at test time. Another classical distinction is made between the inductive and the transductive setting, the latter allowing to use test images (without their annotation) at training time, similarly to semi-supervised learning.

\begin{figure}
    \centering
    \includegraphics[scale=0.35]{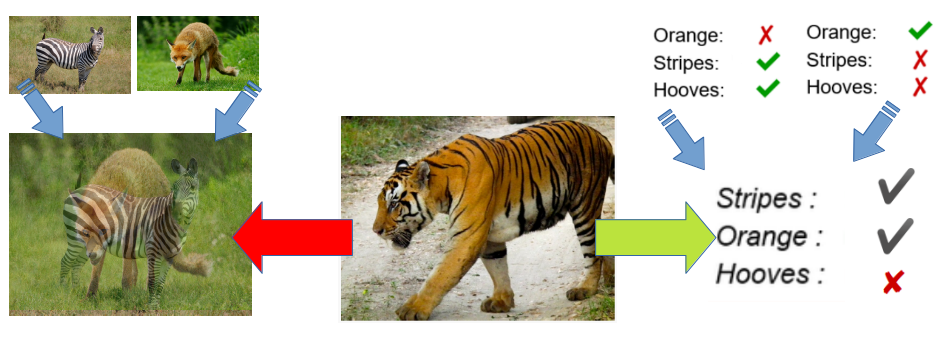}
    \caption{An image of \textit{tiger} differs from a mix of an image of \textit{zebra} and \textit{fox}. At the opposite, learning to discriminate a mix of semantic attributes (or description) of a fox and a zebra makes the proposed model able to better identify a tiger at test time.}
    \label{fig:visual_abstract}
\end{figure}

Recent approaches to ZSL use generative models to produce visual samples from unseen classes based on their semantic descriptions~\cite{Bucher2017,VermaR17,aroraCVPR2018,XianCVPR2018}. With such synthetic samples, we thus have a classical supervised learning setting for the unseen classes as well. One of the most performing approaches in this vein is f-VAEGAN-D2~\cite{XianCVPR2019} that shares the weights of the decoder of a variational autoencoder (VAE) with those of the generator of a generative adversarial network (GAN). Since it is trained in combination with a conditional encoder (and either a conditional or a non-conditional discriminator), it is able to benefit from unlabeled unseen visual samples (transductive setting) to synthesize discriminative image features of unseen classes. It also obtains very good performance in inductive setting as well. 
Narayan \textit{et al.} proposed TF-VAEGAN~\cite{narayan2020latent} that add a decoder which reconstructs the semantic prototypes and a feedback loop from this module to the GAN generator to refine the generated features during both the training and feature synthesis stages. The motivation to add a decoder is that it can provide a complementary information along with the generator, since the latter maps a single prototype to many possible visual instance while the decoder does the opposite. Both pieces of information can be used at test time to create features that are used to learn ZSL and GZSL classifiers.

Beyond the usefulness of the feedback loop, one can see TF-VAEGAN as similar to f-VAEGAN-D2 with an additional loss that regularizes the generator. However, this loss essentially addresses a reconstruction task, while ZSL consists first and foremost in discriminating classes. We thus argue that a loss that regularizes in a way that favor class disambiguation would be more relevant. The loss we propose (Section~\ref{sec:method}) can be integrated to any generative model for ZSL. The idea is to train the generator to learn some ambiguous semantic prototypes built by mixing real available ones, and recognize the corresponding ambiguous classes. This idea may seem similar to the one proposed by Chou \textit{et al.}~\cite{chouICLR2021} who applied mixup~\cite{mixupICLR2017} to the ZSL task, but it has a crucial difference. In fact, when they apply mixup, Chou \textit{et al}. matches the virtual prototype to a corresponding virtual visual feature. Our approach focuses on recognizing a virtual class, thus its label only. Hence, the regularization forces the generator to synthesize discriminative features from unknown class prototypes, some of them being potentially close to some prototypes of the unseen classes (see Section~\ref{sec:influence_subset} for a detailed discussion and evaluation with regards to the availability of the seen/unseen prototypes during training). The generator nevertheless learns without visual samples from seen classes and is thus not constrained by particular images that are not relevant. Indeed, mixing two particular images does not usually result into a meaningful image, while mixing semantic description or attributes may make sense (Fig~\ref{fig:visual_abstract}). Therefore, restricting mixup to the semantic space corresponds better to what is expected in the (G)ZSL task. Our approach is computationally less expensive and leads to better results than Chou \textit{et al.} in practice (Section~\ref{sec:experiment}). Last, the loss they defined is mainly useful in an inductive setting, while the approach we propose can be even more useful in a transductive one.

Our main contribution consists in a regularization loss, that can be applied to any conditional generative-based ZSL model. Integrated to f-VAEGAN-D2 or TFVAEGAN, it improves significantly their performance on different benchmarks, either in inductive or transductive setting.

\section{State of the Art} \label{sec:sota}
Early approaches in ZSL relied on attribute prediction~\cite{lampert2009}, ridge regression~\cite{romera2015eszsl} or triplet-loss~\cite{frome2013devise,lecacheux2019tripletloss,lecacheux2019classical}. We refer to~\cite{lecacheux2021zsl_sota} for a detailed overview of these approaches as we focus on generative approaches in the following.

In order to address the biased prediction towards seen classes, generative approaches synthesize visual features of unseen classes from their semantic features with generative models like Variational Autoencoders (VAEs) or Generative adversarial networks (GANs). Xian \textit{et al.}~\cite{XianCVPR2018} combines a conditional Wasserstein GAN~\cite{arjovsky2017wasserstein} with a categorization network to generate more discriminative features. Bucher \textit{et al}.~\cite{Bucher2017} proposed three different conditional GANs to generate features, Generative Moment Matching Network (GMMN), AC-GAN, and Denoising Auto-Encoder. Other works use conditional VAE. Arora \textit{et al.}~\cite{aroraCVPR2018} integrates an attribute regressor and a feedback mechanism into a VAE-based model to generate more discriminative features and ensure that the generated features are semantically close to the distribution of the real features. Schonfeld \textit{et al. }~\cite{schonfeldCVPR2019} proposes to align the visual features and the corresponding semantic embeddings in a shared latent space, using two Variational Autoencoders (VAEs). Recent works take advantage of both GANs and VAEs by combining them with shared decoder and generator. Xian \textit{et al}.~\cite{XianCVPR2019} proposes a VAEGAN-based model that leverages the unlabeled instances under the transductive setting via an additional unconditional discriminator. Similar to the idea proposed by Arora \textit{et al.}, Narayan \textit{et al.}~\cite{narayan2020latent} augments the \texttt{f-VAEGAN-D2} method with a semantic embedding decoder and a feedback mechanism to enforce a semantic consistency and improve feature synthesis. In this work, we propose to enrich a conditional VAEGAN-based ZSL method with an auxiliary task as well while focusing on another aspect, more related to the ability to discriminate classes, namely reducing ambiguities among categories. Such a goal can be useful beyond zero-shot learning, for tasks that aim at relating ambiguous visual and semantic information such as multimodal entity linking~\cite{adjali2020ecir} and retrieval~\cite{myoupo2010lncs,znaidia2012icmr}, cross-modal retrieval~\cite{tran15icmr,tran16cvpr,chami17icmr} or classification~\cite{tran16ivl}. 

To alleviate the domain shift problem, transductive ZSL methods are proposed to leverage the unlabeled unseen-class data at training. Xian \textit{et al}.~\cite{xian2017} and Ye \textit{et al}.~\cite{YeCVPR2017} proposed to use graph-based label propagation while Verma \textit{et al}.~\cite{VermaR17} uses an Expectation-Maximization (EM) strategy, where pseudo-labeled unseen class examples were used to update the parameter estimates of unseen class distributions. Generative models  were also applied to transductive ZSL. Paul \textit{et al}.~\cite{PaulCVPR2019} leverages Wasserstein GAN to synthesize the unseen domain distribution via minimizing the marginal difference between the true latent space representation of the unlabeled samples of unseen classes and the synthesized space. f-VAEGAN-D2 and TF-VAEGAN can also be applied to transductive settings. 

\begin{figure*}[bt]
    \centering
    \includegraphics[width=0.9\textwidth]{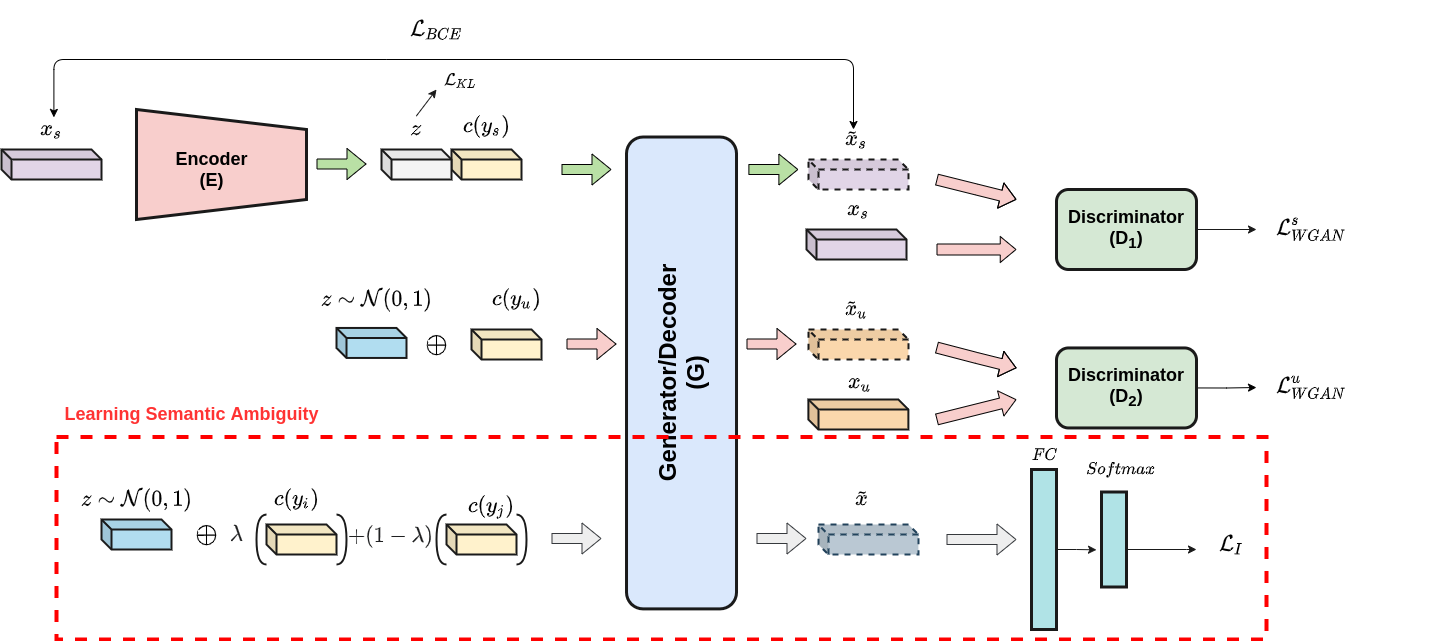}
    \caption{Our contribution, highlighted by the red dashed rectangle, included to a model similar to f-VAEGAN-D2~\cite{XianCVPR2019}. The encoder $E$ takes the real seen features $x_s$ as input and outputs a latent code $z$, which is then input together with embeddings $c(y^s)$ to the generator $G$ that synthesizes features $\tilde{x}_s$. The generator $G$ synthesizes features of unseen classes from the class prototypes $c(y^u)$ concatenated with random noise $z$. The two discriminators $D_1$ and $D_2$ learn to distinguish between real and synthesized features. We introduce a novel task to train the generator $G$ : First, virtual \textit{ambiguous} classes are constructed as convex combinations of \textit{real} classes, then, the generator $G$ synthesizes a feature $\tilde{x}$ from the corresponding \textit{ambiguous} class prototypes concatenated with random noise. Further, the synthesized features $\tilde{x}$ are used to perform a classification task.}
    \label{fig:archi_overview}
\end{figure*}

Since its advent, interpolation-based regularization has been shown to be a surprisingly effective method to improve generalization and robustess on both supervised and semi-supervised settings. Zhang \textit{et al.} \cite{mixupICLR2017} proposed \textit{mixup}, a data augmentation technique for image classification based on the Vicinal Risk Minimization principle \cite{ChapelleNIPS2000}. It consists in training models on virtual examples constructed as the convex combinations of pairs of visual data samples and their corresponding labels. 
%
This simple approach has shown to be an effective model regularizer that favor linear behavior in-between training examples. Recently, Chou \textit{et al.} \cite{chouICLR2021} applied \textit{mixup} to zero-shot learning. Similarly to our method, they interpolate both the visual samples and the semantic prototypes. However, unlike our approach, they used mixup as a direct data augmentation approach, while we apply the interpolation in the conditional space of a generative ZSL model and propose a specific regularization loss in the semantic space. More specifically, we train a conditional generative ZSL model to recognize virtual ambiguous classes. The generator synthesizes features from the corresponding ambiguous class prototypes, which are then used to perform the classification task. In practice the difference between the linear interpolation we propose and the usual mixup setting used by~\cite{chouICLR2021} is reflected by the mixing proportion leading to the best performances. Indeed, as an augmentation data approach, mixup usually have better performances with a mixing proportion that must be either close to 0 or 1, making the new virtual samples pretty close to the original ones. In contrast, we obtain the best performances with a mixing proportion close to 0.5, making the new \textit{virtual} classes completely distinct from the real onese. These classes are different from the actual unseen classes used at test time, but allow the generator to be regularized in some `empty' parts of the semantic space.


\section{Method} \label{sec:method}
\subsection{Problem setting and notation}
Let us consider a set of images $X = \{x_1,..., x_l, x_{l+1},..., x_t\}$ encoded in the image feature space $\mathcal{X} = \mathbb{R}^d$ and two disjoint sets of class labels: a seen class label set $\mathcal{Y}^s$ and the unseen one $\mathcal{Y}^u$. The set of class prototypes is denoted as $C = \{ c(y) \|  y \in \mathcal{Y}^s \cup \mathcal{Y}^u, c(y) \in \mathcal{C} \}$. Usually, $c(y)$ is a vector of binary attributes, but may be word embeddings when one wants to describe a large set of classes~\cite{frome2013devise,lecacheux2020webly,lecacheux2020using}. The first $l$ samples $x_s$, with $s \leq l$, are labeled samples from seen classes $y^s \in \mathcal{Y}^s$ and the remaining samples $x_u$, with $ l+1 \leq u \leq t$, are unlabeled data from novel classes, $y^u \in \mathcal{Y}^u$. 
In the inductive setting, the training set contains only labeled seen classes examples, and the semantic information about both seen and unseen classes. 
%
In the transductive setting, the training set contains both labeled (seen classes) and unlabeled (unseen classes) data samples.
In fact, there is an ambiguity in the definition of the \textit{transductive} setting in the literature, as there is more than one definition of this setting. Indeed, Le Cacheux \textit{et al.}~\cite{lecacheux2021zsl_sota} defines the \textit{class-transductive} setting, in which class prototypes of both seen and
unseen classes are available during the training phase, and the \textit{instance-transductive} setting, where both prototype and unlabeled images from unseen classes are available. The class-transductive setting is sometimes referred as inductive, as the author considers that the unseen prototypes need to be available to generate the unseen visual samples used to learn the classifiers. In this paper, we refer to ``transductive'' setting when unseen prototypes are used for any other usage than generating unseen visual features.

In zero-shot learning, the goal is to predict the label of images that belong to unseen classes, i.e. $f_{z s l}: \mathcal{X} \rightarrow \mathcal{Y}^{u}$, while in the GZSL scenario, the task is to predict labels of images that can belong to either seen or unseen classes, i.e. $f_{g z s l}: \mathcal{X} \rightarrow \mathcal{Y}^{s} \cup \mathcal{Y}^{u}$, with $f$ a compatibility function that computes the likelihood of an image to belong to a class.

\subsection{Learning Semantic Ambiguities}
We consider a generative approach to ZSL, for which a conditional model $G(.)$ is trained with visual samples $x^s$ and prototypes $c(y^s)$ from seen classes $y^s$. At test time, it is able to generate visual samples $x^u$ from a prototype $c(y^u)$ for unseen classes $y^u$, which are used to train a classifier in a fully supervised fashion. Applying mixup to this setting such as \cite{chouICLR2021} consists in augmenting the training data with virtual pairs:
\begin{equation}\label{eq:chou_model}
  \begin{array}{rl}
     \tilde{c}=&\lambda c(y^s_i)+(1-\lambda) c(y^s_j) \\
     \tilde{x}=&\lambda x^s_{i}+(1-\lambda) x^s_{j} 
   \end{array}
 \end{equation}
 where $\lambda$ is a hyperparameter to determine and $((x_i,y_i),(x_j,y_j))$ is a couple of annotated data from seen classes randomly selected. As explained above, we argue that using the visual samples biases the generator towards seen classes. Such a bias was identified for the former ZSL approaches by Xian \textit{et al.}~\cite{xian2017} and led to the definition of generalized zero-shot learning (GZSL), that is the most common and challenging setting in the literature.

 We thus adopt a different strategy: focusing on learning the ambiguities in the semantic space only. Indeed, as illustrated in Fig~\ref{fig:visual_abstract} mixing two particular images does not usually make sense, because of strong inconsistencies at the pixel level. In contrast, mixing semantic information \textit{may} make sense. First, it is at the origin of a large bestiary in fantasy literature, in science fiction and in heroic fantasy. A unicorn is described semantically as a horse with a corn, and no picture of such creature has been taken up to date, although many artist proposed some visual representation of it. However, we do not expect the generator to produce features that would result into a plausible representation but rather features able to emphasis the differences between classes, in order to better distinguish them. Hence, we propose to regularize the model with such a constraint only, while being independent of any particular existing visual representation.
 In practice, we create ambiguous classes as a linear interpolation of real semantic prototype pairs and their labels:

\begin{equation}\label{eq:our_mix}
  \begin{aligned}
        \tilde{c}&=\lambda c(y_i)+(1-\lambda) c(y_j) \\
        \tilde{y}&=\lambda y_{i}+(1-\lambda) y_{j}
  \end{aligned}
\end{equation}
The hyperparameter $\lambda$ can be a fixed value or more generally a random variable $\lambda\sim\Lambda$. During the learning phase, the generator $G$ synthesizes  feature $\hat{x} \in \mathcal{X}$ from a latent code $z \sim \mathcal{N}(0,1)$ conditioned by the {ambiguous} class prototype $\tilde{c} \in \mathcal{C}$. This image is then used as input to a classifier $f$, leading to the proposed regularization loss as:
\begin{equation}
\label{eq:loss_mix}
    \mathcal{L}_I = \mathbb{E}_{z,\lambda} [ l(f (\hat{x}),\tilde{y})]
\end{equation}
where $l$ is the cross-entropy between the input $\hat{x}=G([z;\tilde{c}])$ and the target $\tilde{y}$. 
With its generic formulation, the proposed regularization can be applied to a large number of generative models. In the following, we integrate it to f-VAEGAN-D2~\cite{XianCVPR2019} as illustrated in Fig.~\ref{fig:archi_overview} by adding $\mathcal{L}_I$ to their losses. In that case, the total loss to minimize is $\mathcal{L}=\mathcal{L}_{BCE}+\gamma\mathcal{L}_{WGAN}^s+\mathcal{L}_{WGAN}^u+\mathcal{L}_I$ with $\gamma$ an hyperparameter (see section~\ref{sec:impl_details}). It can nevertheless be added to any generative-based ZSL model, and we show how it performs with TFVAEGAN in section~\ref{sec:with_tfvaegan}.

\section{Experimental Evaluation} \label{sec:experiment}
\subsection{Datasets and Metrics}
We evaluate our method on four datasets that are commonly used in the ZSL literature, namely Caltech UCSD Birds 200-2011 (CUB)~\cite{WahCUB_200_2011}, SUN Attribute dataset~\cite{Patterson2012SunAttributes}, Oxford Flowers (FLO)~\cite{Nilsback08} and Animals with Attributes (AWA2) \cite{xian2017}. Their main characteristics are reported in Table~\ref{tab:datasets}

\begin{table*}
  \centering
  \begin{tabular}{c|cccc}
  \textbf{Dataset} & \textbf{Type} & \textbf{\# images} & \textbf{\#classes} & \textbf{\# attributes} \\ 
  \hline
  CUB \cite{WahCUB_200_2011} & Fine & 11,788 & 200 & 1024 \\
  SUN \cite{Patterson2012SunAttributes} & Fine & 14,340 & 717 & 102 \\
  FLO \cite{Nilsback08} & Fine & 8,189 & 102 & 1024 \\
  AwA2 \cite{xian2017} & Coarse & 37,322 & 50 & 85
  \end{tabular}%
  \caption{Main characteristics of the dataset used. The granularity stands for  fine-grained or coarse-grained classification}
  \label{tab:datasets}
\end{table*}
  
We applied the evaluation protocol of Xian \textit{et al.}~\cite{xian2017}, relying on the ``proposed splits'' that insure that none of the test classes appear in ImageNet, since it is used to pre-train the visual feature extractor. The performances are reported in terms of average per-class top-1 accuracy (T1) for ZSL settings, and with the harmonic mean (\textbf{H}) of the average per-class top-1 accuracy on seen (s) and unseen (u) classes for GZSL. Unless otherwise specified, we use 2048-dimension 101-ResNet features as visual embeddings for all the datasets. For class semantic prototypes of CUB and FLO, we adopt the 1024-dim sentence embeddings of character-based CNN-RNN model generated from fine-grained visual descriptions \cite{reed2016}. For AWA2, the binary attributes relate to e.g animal species (\textit{``fish, bird, plankton''}), color (\textit{``black, brown, blue''}), behaviour (\textit{``hibernate, timid, slow''}) and other features. For SUN, they rather relate to function/affordances, materials, spatial envelope and surface properties.
We compare our method to \texttt{TF-VAEGAN} \cite{narayan2020latent}, \texttt{f-VAEGAN-D2} \cite{XianCVPR2019}, \texttt{CLSWGAN} \cite{XianCVPR2018} and \texttt{LisGAN} \cite{LiCVPR2019}. 

To compare the methods over several benchmarks and estimate their aggregated merit, we adopt the median normalized relative gain (mNRG)~\cite{tamaazousti19pami}. Indeed, such a comparison can be biased if one uses a simple average over different benchmarks. mNRG exhibits several interesting features such as an independence to outlier scores, coherent aggregation or time consistency. Its main drawback is that a reference method has to be chosen, from which the performance of each method is measured, according to a unique aggregated score, possibly negative if the method performs globally worse. In our case, we choose CLSWGAN~\cite{XianCVPR2018} in inductive settings as reference. For the comparison with fine-tuned features, we use f-VAEGAN-D2 in inductive settings as reference. We compute the  mNRG by aggregating the accuracy for ZSL and the harmonic mean $H$ of seen and unseen accuracy for GZSL. By definition, the score of the reference is 0. If $mNRG<0$ then the method performs globally worse than the references over all datasets. When $mNRG=1$ the method obtains the best performances on a majority of datasets (at least 3 on 4 in our experiments).

\subsection{Implementation details}\label{sec:impl_details}
The generator $G$ and discriminators $D_1$ and $D_2$ are implemented as two-layer fully connected networks with 4096 hidden units. The generator is updated every 5 discriminator iterations~\cite{arjovsky2017wasserstein}. The function $f$ used in equation (\ref{eq:loss_mix}) is implemented as a two-layers fully connected network that takes an input synthesized feature of size $d=2048$, has a hidden layer of size $4096$ and outputs a probability distribution with regards to all classes of interest. We use LeakyReLU activation everywhere, except at the output of $G$, where a sigmoid non-linearity is applied before the binary cross-entropy loss $\mathcal{L}_{BCE}$. ZSL and GZSL classifiers are implemented as a single layer perceptron of size $2,048$, trained for 20 epochs. We use Adam optimizer with a learning rate of 0.0001. 
Our (PyTorch) code is based on the one of \cite{narayan2020latent} and is available at \url{https://github.com/hanouticelina/lsa-zsl}. We determined that the hyperparameters $\gamma=10$ and the gradient penalty of the WGAN loss $\lambda_{WGAN}=10$ allowed us to obtain similar performances as those reported in \cite{XianCVPR2019,narayan2020latent}, although they are sometimes different to the hyperparameter values reported in these papers. Using the code of \cite{narayan2020latent}, it is possible to reproduce their experiments and those of \cite{XianCVPR2019} for the inductive setting only. Our code allows to reproduce the experiments under the transductive setting as well.

\begin{table*}[tb]
  \centering
  \resizebox{0.5\textwidth}{!}{%
  \begin{tabular}{ll|llll|l}
  \multicolumn{2}{l}{\multirow{3}{*}{}} & \multicolumn{5}{c}{\textbf{Zero-shot Learning}}     \\
  \multicolumn{2}{l}{}  & \textbf{CUB}  & \textbf{FLO}   & \textbf{SUN}   & \multicolumn{1}{l}{\textbf{AWA2}} &     \\
  \multicolumn{2}{l}{}    & \multicolumn{1}{c}{\textbf{T1}} & \multicolumn{1}{c}{\textbf{T1}}  & \multicolumn{1}{c}{\textbf{T1}} & \multicolumn{1}{c|}{\textbf{T1}}  & mNRG \\
  \hline
  IN & \texttt{CLSWGAN} \cite{XianCVPR2018} & 57.3 & 67.2 & 60.8 & 68.2 & 0 [ref]\\
     & \texttt{LisGAN} \cite{LiCVPR2019} & 58.8 & 69.6 & 61.7 & 70.6 & 2.0\\
     & \texttt{f-VAEGAN-D2} (*) \cite{XianCVPR2019} & 61.0 & 67.7 & 64.7 & 71.1 & 3.3\\
     & \texttt{TF-VAEGAN} (*) \cite{narayan2020latent} & 63.2 & \textbf{70.4} & 64.3 & \textbf{73.2} & \textbf{4.2}\\
     & \textbf{Ours } & \textbf{70.7} & 69.2 & \textbf{64.7} & 71.9 & 3.8\\
  \hline
  TR & \texttt{ALE-trans}  \cite{xian2017} & 54.5 & 48.3 & 55.7 & 70.7 & -4.0\\
     & \texttt{GFZSL} \cite{VermaR17} & 50.0 & 85.4 & 64.0 & 78.6 & 6.8\\
     & \texttt{DSRL} \cite{YeCVPR2017} & 48.7 & 57.7 & 56.8 & 72.8 & -6.3\\
     & \texttt{f-VAEGAN-D2} (*) \cite{XianCVPR2019} & 74.2 & 89.1 & 70.1 & 89.8 & 19.3\\
     & \texttt{TF-VAEGAN} (*) \cite{narayan2020latent} & 77.2& \textbf{92.6} & 70.1& 92.1 & 21.9\\
     & \textbf{Ours} & \textbf{80.6}  & 89.3 & \textbf{71.7}& \textbf{92.8} & \textbf{22.7}\\
  \hline
  \multicolumn{2}{l}{\multirow{3}{*}{}} & \multicolumn{5}{c}{\textbf{ZSL with fine-tuned features}}     \\
  \hline
  FT-IN & \texttt{f-VAEGAN-D2 (*) } \cite{XianCVPR2019} & 74.1 & 70.5 & \textbf{64.5} & 69.9 & 0 [ref]\\
        & \texttt{TF-VAEGAN (*) } \cite{narayan2020latent} & 72.5 & 70.6 & 64.1 & 68.5 & -0.9\\
        & \textbf{\textbf{Ours}} & \textbf{83.3} & \textbf{72.8} & 64.0 & \textbf{70.4} & \textbf{1.4}\\
  \hline
  FT-TR & \texttt{f-VAEGAN-D2 (*) } \cite{XianCVPR2019} & 82.1 & 95.6 & 68.5 & 89.9 & 14.0\\
        & \texttt{TF-VAEGAN (*) } \cite{narayan2020latent} & 85.1 & \textbf{96.0} & \textbf{73.8} & \textbf{93.0} & \textbf{17.1}\\
        & \textbf{\textbf{\textbf{\textbf{Ours}}}} & \textbf{86.1} & 95.8 & 70.0 & 91.1 & 16.6\\
  \end{tabular}
  }
  \caption{\textbf{State-of-the-art comparison} Accuracy for ZSL on the “proposed split” of \cite{xian2017}, both inductive (IN) and transductive (TR) results are shown. \textit{Avg} is the average score over the four datasets. Models marked with * were partially re-implemented.}
  \label{tab:zsl-sota}
  \end{table*}
  
  \begin{table*}[tb]
  \centering
  \resizebox{\textwidth}{!}{%
  \begin{tabular}{lllllllllllllll}
  \multicolumn{2}{l}{\multirow{3}{*}{}} &
    \multicolumn{12}{c}{\textbf{Generalized Zero-shot Learning}} &
     \\
  \multicolumn{2}{l}{} &
    \multicolumn{3}{c}{\textbf{CUB}} &
    \multicolumn{3}{c}{\textbf{FLO}} &
    \multicolumn{3}{c}{\textbf{SUN}} &
    \multicolumn{3}{c}{\textbf{AWA2}} &
     \\
  \multicolumn{2}{l}{} &
    \multicolumn{1}{c}{u} &
    \multicolumn{1}{c}{s} &
    \multicolumn{1}{c|}{\textbf{H}} &
    \multicolumn{1}{c}{u} &
    \multicolumn{1}{c}{s} &
    \multicolumn{1}{c|}{\textbf{H}} &
    \multicolumn{1}{c}{u} &
    \multicolumn{1}{c}{s} &
    \multicolumn{1}{c|}{\textbf{H}} &
    \multicolumn{1}{c}{u} &
    \multicolumn{1}{c}{s} &
    \multicolumn{1}{l|}{\textbf{H}} &
    \textbf{$mNRG$} \\ \hline
  IN &
    \multicolumn{1}{l|}{\texttt{CLSWGAN} \cite{XianCVPR2018}} &
    43.7 &
    57.7 &
    \multicolumn{1}{l|}{49.7} &
    59.0 &
    73.8 &
    \multicolumn{1}{l|}{65.6} &
    42.6 &
    36.6 &
    \multicolumn{1}{l|}{39.4} &
    57.9 &
    61.4 &
    \multicolumn{1}{l|}{59.6} &
    0 [ref] \\
   &
    \multicolumn{1}{l|}{\texttt{LisGAN} \cite{LiCVPR2019}} &  46.5 &   57.9 &   \multicolumn{1}{l|}{51.6} &  57.7 &  \textbf{83.8} & \multicolumn{1}{l|}{68.3} &  42.9 &  37.8 &  \multicolumn{1}{l|}{40.2} &  52.6 &  76.3 &  \multicolumn{1}{l|}{62.3} & 2.3 (1.9) \\ 
   & 
    \multicolumn{1}{l|}{\texttt{f-VAEGAN-D2} (*) \cite{XianCVPR2019}} &  48.5 &  60.2 &  \multicolumn{1}{l|}{53.7} &  56.8 &  73.9 &  \multicolumn{1}{l|}{64.2} &  45.1 &  38.0 &   \multicolumn{1}{l|}{41.3} &  \textbf{57.6} &   70.6 &  \multicolumn{1}{l|}{63.5} &  2.9 (3.9) \\ 
   &
    \multicolumn{1}{l|}{\texttt{TF-VAEGAN} (*) \cite{narayan2020latent}} &  52.2 &  62.7 &
    \multicolumn{1}{l|}{56.9} &  62.4 &  83.5 &
    \multicolumn{1}{l|}{\textbf{71.4}} &  41.3 &  \textbf{39.2} &  \multicolumn{1}{l|}{40.2} &
    52.8 &  \textbf{81.9} &  \multicolumn{1}{l|}{64.2} &  5.2 (4.6) \\  
   &
    \multicolumn{1}{l|}{\cite{chouICLR2021}} &   41.4 & 49.7 & \multicolumn{1}{l|}{45.2}&   - & -  & \multicolumn{1}{l|}{-} &   29.9 & 40.2& \multicolumn{1}{l|}{34.3} &   65.1 & 78.9 & \multicolumn{1}{l|}{71.3}
     & xx (-4.5) \\
   &
    \multicolumn{1}{l|}{\textbf{Ours}} &
    \textbf{60.3} &
    \textbf{75.9} &
    \multicolumn{1}{l|}{\textbf{67.2}} &
    \textbf{62.6} &
    81.5 &
    \multicolumn{1}{l|}{70.8} &
    \textbf{45.2} &
    39.0 &
    \multicolumn{1}{l|}{\textbf{41.8}} &
    \textbf{57.6} &
    76.4 &
    \multicolumn{1}{l|}{\textbf{65.6}} &
    \textbf{5.6} (6)\\ \hline
  TR &
    \multicolumn{1}{l|}{\texttt{ALE-trans} \cite{xian2017}} &  23.5 &  45.1 &  \multicolumn{1}{l|}{30.9} &  13.6 &  61.4 &  \multicolumn{1}{l|}{22.2} &  19.9 &  22.6 &  \multicolumn{1}{l|}{21.2} &  12.6 &  73.0 &  \multicolumn{1}{l|}{21.5} &  -28.5 \\ 
   &
    \multicolumn{1}{l|}{\texttt{GFZSL} \cite{VermaR17}} &
    24.9 & 45.8 &  \multicolumn{1}{l|}{32.2} &  21.8 & 75.0 &  \multicolumn{1}{l|}{33.8} &  0.0 & 41.6 &  \multicolumn{1}{l|}{0.0} &  31.7 &  67.2 & \multicolumn{1}{l|}{43.1} &  -24.7 \\   
   &
    \multicolumn{1}{l|}{\texttt{DSRL} \cite{YeCVPR2017}} &  17.3 & 39.0 &\multicolumn{1}{l|}{24.0} &  26.9 & 64.3 &\multicolumn{1}{l|}{37.9} &  17.7 &  25.0 &  \multicolumn{1}{l|}{20.7} &  20.8 &  74.7 &  \multicolumn{1}{l|}{32.6} &  -26.4 \\ 
   &
    \multicolumn{1}{c|}{\texttt{f-VAEGAN-D2} (*) \cite{XianCVPR2019}} &
    \multicolumn{1}{c}{65.6} &
    \multicolumn{1}{c}{68.1} &
    \multicolumn{1}{c|}{66.8} &
    \multicolumn{1}{c}{78.7} &
    \multicolumn{1}{c}{87.2} &
    \multicolumn{1}{c|}{82.7} &
    \multicolumn{1}{c}{60.6} &
    \multicolumn{1}{c}{41.9} &
    \multicolumn{1}{c|}{49.6} &
    84.8 &
    88.6 &
    \multicolumn{1}{l|}{86.7} &
    17.1 \\ 
   &
    \multicolumn{1}{l|}{\texttt{TF-VAEGAN} (*) \cite{narayan2020latent}} &
    \multicolumn{1}{c}{69.1} &
    \multicolumn{1}{c}{\textbf{75.1}} &
    \multicolumn{1}{c|}{72.0} &
    \multicolumn{1}{c}{83.8} &
    \multicolumn{1}{c}{91.9} &
    \multicolumn{1}{c|}{87.6} &
    \multicolumn{1}{c}{\textbf{62.5}} &
    \multicolumn{1}{c}{\textbf{46.8}} &
    \multicolumn{1}{c|}{\textbf{53.5}} &
    84.5 &
    \textbf{90.2} &
    \multicolumn{1}{l|}{87.2} &
    22.2 \\
   &
    \multicolumn{1}{l|}{\textbf{Ours}} &
    \textbf{74.2} &  70.5 &  \multicolumn{1}{l|}{\textbf{72.3}} &  \textbf{85.1} &
    \textbf{92.2} &  \multicolumn{1}{l|}{\textbf{88.5}} &
    59.5 &  46.0 &  \multicolumn{1}{l|}{51.8} &  \textbf{86.7} &  88.5 &  \multicolumn{1}{l|}{\textbf{87.6}} &  \textbf{22.8} \\ 
    \hline
  \multicolumn{2}{l}{\multirow{3}{*}{}} &
    \multicolumn{12}{c}{\textbf{GZSL with fine-tuned features}} &
     \\
     \hline
  FT-IN & \multicolumn{1}{l|}{\texttt{f-VAEGAN-D2} \cite{XianCVPR2019}} & 70.4 & 79.2 &\multicolumn{1}{l|}{74.5}                            & 59.0 & \textbf{93.8} & \multicolumn{1}{l|}{72.4} & \textbf{50.6} & 37.4 & \multicolumn{1}{l|}{\textbf{43.0}} &  60.1 & 79.3 & \multicolumn{1}{l|}{68.3} &
    0 [ref] \\
   & \multicolumn{1}{l|}{\texttt{TF-VAEGAN} \cite{narayan2020latent}}   &
   64.1 &  79.7 &
   \multicolumn{1}{l|}{71.0} &
   \textbf{64.6} &
   90.3 & \multicolumn{1}{l|}{\textbf{75.3}}  &
   38.2 & \textbf{43.8} & \multicolumn{1}{l|}{40.8} &
   57.8 & \textbf{87.3} & \multicolumn{1}{l|}{\textbf{69.5}}  &
   -0.5 (-2.2)\\
   &
   \multicolumn{1}{l|}{\cite{chouICLR2021}} & 69.2 &76.4& \multicolumn{1}{l|}{72.6} & -& -& \multicolumn{1}{l|}{-}&50.5 & 43.1& \multicolumn{1}{l|}{46.5}&69.0 & 86.5& \multicolumn{1}{l|}{76.8} & xx (3.5) \\
   & 
   \multicolumn{1}{l|}{\textbf{Ours}}  & \textbf{65.6} &
   \textbf{82.7} &
   \multicolumn{1}{l|}{\textbf{73.2}} & 
   62.0 & 89.4 & \multicolumn{1}{l|}{73.2} & 43.9 & 41.8 & \multicolumn{1}{l|}{42.8} &  \textbf{60.7}  &
   80.1 & \multicolumn{1}{l|}{69.0} &    
   \textbf{0.25} (-0.2) \\
   \hline
  FT-TR & \multicolumn{1}{l|}{\texttt{f-VAEGAN-D2} (*) \cite{XianCVPR2019}}   & 77.1 &82.0 &\multicolumn{1}{l|}{79.5} & 92.7&97.4&\multicolumn{1}{l|}{94.9} & \textbf{62.0}& 38.6 & \multicolumn{1}{l|}{47.6} & 83.9&\textbf{95.5} & \multicolumn{1}{l|}{\textbf{89.3}} & 12.3 \\ 
  & \multicolumn{1}{l|}{\texttt{TF-VAEGAN} (*) \cite{narayan2020latent}} & 80.1 & \textbf{80.0} &\multicolumn{1}{l|}{80.0} &82.7 &94.3 &\multicolumn{1}{l|}{88.2} &60.5 &43.6 &\multicolumn{1}{l|}{\textbf{50.7}} & 79.7 & 89.6                  & \multicolumn{1}{l|}{84.4} & 11.8 \\
  & \multicolumn{1}{l|}{\textbf{Ours}} & \textbf{81.2} & 82.6 & \multicolumn{1}{l|}{\textbf{81.9}}                   & \textbf{93.0} & \textbf{97.7} & \multicolumn{1}{l|}{\textbf{95.3}}& 47.5 & \textbf{48.6} & \multicolumn{1}{l|}{48.0} & \textbf{85.6} & 91.3 & \multicolumn{1}{l|}{88.3} &\textbf{13.7} 
  \end{tabular}%
  }
  \caption{\textbf{State-of-the-art comparison} Accuracy for GZSL on the “proposed split” of \cite{xian2017}, both inductive (IN) and transductive (TR) results are shown. We measure Top-1 accuracy on seen (s) and unseen (u) classes as well as their harmonic mean (H). \textbf{$\tilde{H}$} is the average over the four datasets. Models marked with * were partially re-implemented. The mNRG in parentheses is computed without the score on FLO.}
  \label{tab:gzsl-sota}
  \end{table*}
\subsection{State-of-the-art comparison}\label{sec:sota-comparison}
Table \ref{tab:zsl-sota} and Table \ref{tab:gzsl-sota} show the comparison to the state-of-the-art. For inductive ZSL setting, our model performs globally better than all other methods with the highest mNRG score. It also achieves the best score on CUB and SUN. In the transductive ZSL setting, our approach obtains a mNRG score of 22.8, establishing a new transductive ZSL state-of-the-art on CUB, SUN and AWA2. The comparison to \cite{chouICLR2021} is particular since they report results on three of the considered datasets only. Without fine-tuning, their results is far above other methods on AWA2 but also far below on CUB and SUN. 

Unsurprinsingly, in the GZSL setting, feature generating approaches 
obtain better results than others. We also note that the accuracy on unseen classes ($u$) and the one on seen classes ($s$) are better balanced. Our model outperforms the existing methods for both inductive and transductive GZSL settings. In particular, in the inductive GZSL setting, our model obtains 67.2 $\%$ on CUB, significantly improving those obtained previously (56.9$\%$). By reducing the bias towards seen classes, we globally achieve better performance on unseen classes. However, the scores on seen classes may slightly decrease, in particular in inductive setting. It is nevertheless compensated by the gain on the unseen classes.

We also conducted some experiments with fine-tuned features, wit the same features as those used in \cite{XianCVPR2019, narayan2020latent}. To compute the global score mNRG for this experiment, we used f-VAEGAN-D2 in transductive setting as a baseline.  In the ZSL setting, the results of TF-VAEGAN are globally better than ours in the transductive setting, both being significantly above f-VAEGAN. However, in the inductive setting, the results of TF-VAEGAN are below the baseline while ours are still slightly above.

In the GZSL settings, TF-VAEGAN still has a lower mNRG score than the baseline f-VAEGAN-D2 in the inductive setting and quite comparable score in the transductive ones. Our approach obtains performances in line with f-VAEGAN in the inductive but significantly outperforms the two other approaches in the transductive settings when compared over the four datasets.

\subsection{Ablation study}

\begin{table*}[tb]
\resizebox{\textwidth}{!}{%
\begin{tabular}{ccccc|cccccccccccc}
\multicolumn{1}{l}{}  & \multicolumn{4}{c|}{\textbf{Zero-shot Learning}} & \multicolumn{12}{c}{\textbf{Generalized Zero-shot Learning}}   \\
  & \textbf{CUB} & \textbf{FLO} & \textbf{SUN} & \textbf{AWA2} & \multicolumn{3}{c|}{\textbf{CUB}} & \multicolumn{3}{c|}{\textbf{FLO}} & \multicolumn{3}{c|}{\textbf{SUN}}                                                      & \multicolumn{3}{c}{\textbf{AWA2}} \\
  & \textbf{T1} & \textbf{T1} & \textbf{T1} & \textbf{T1} & u & s & \multicolumn{1}{c|}{\textbf{H}} & u & s   & \multicolumn{1}{c|}{\textbf{H}} & u & s  & \multicolumn{1}{c|}{\textbf{H}} & u & s & \textbf{H}   \\ \hline
\multicolumn{1}{c|}{$\lambda \sim$ Beta(0.3, 0.3)}                          & \textbf{80.9} & 72.1 & 71.3 & 87.7 & 74.4 & 70.0 & \multicolumn{1}{c|}{72.1} & 67.3 & 89.7 & \multicolumn{1}{c|}{76.9} & 60.6 & 44.9 & \multicolumn{1}{c|}{51.6} & 79.6 & 86.3 & 82.8 \\
\multicolumn{1}{c|}{$\lambda \sim$ Uniform(0, 1)}                           & \multicolumn{1}{c}{80.8}        & \multicolumn{1}{c}{86.3}        & \multicolumn{1}{c}{\textbf{72.2}}        & 87.0                             & 75.0                     & \multicolumn{1}{c}{69.0} & \multicolumn{1}{c|}{71.9} & 83.1                     & \multicolumn{1}{c}{91.1} & \multicolumn{1}{c|}{86.9} & 60.1 & \multicolumn{1}{c}{45.5} & \multicolumn{1}{c|}{51.8} & 79.1    & 87.5     & 83.1         \\
\multicolumn{1}{c|}{$\lambda \sim \mathcal{N}(0.5, 0.25)$} & 80.7 & \textbf{89.7} & 71.4 & 89.2                    & \multicolumn{1}{l}{74.3} &            69.8 & \multicolumn{1}{l|}{72.0}           & \multicolumn{1}{l}{86.3} & 91.0                     & \multicolumn{1}{l|}{\textbf{88.5}}       & \multicolumn{1}{l}{60.6} & 45.6            & \multicolumn{1}{l|}{\textbf{52.1}}           & 85.9 & 88.7 & 87.2 \\
\multicolumn{1}{c|}{$\lambda = 0.5$}                                                       & 80.6 & 89.3                            & 71.7 & \textbf{92.8} & \multicolumn{1}{l}{74.2} & 70.5 & \multicolumn{1}{l|}{\textbf{72.3}}       & \multicolumn{1}{l}{85.1} & 92.2 & \multicolumn{1}{l|}{\textbf{88.5}} & \multicolumn{1}{l}{59.5} & 46.0 & \multicolumn{1}{l|}{51.8}       & 86.7    & 88.5     & \textbf{87.6}         \\
\multicolumn{1}{c|}{$\lambda = 0.2$}                                                      & \textbf{80.9} & 86.9                            & 71.7 & 88.6 & \multicolumn{1}{l}{74.2}     &             69.8             & \multicolumn{1}{l|}{71.9}           & \multicolumn{1}{l}{83.3} & 92.1                     & \multicolumn{1}{l|}{87.5}       & \multicolumn{1}{l}{59.3}     &  45.3                        & \multicolumn{1}{l|}{51.3}           &     81.2    &   83.6       & 82.4  

\end{tabular}%
}
\caption{Transductive ZSL and GZSL results with different values of the $\lambda$ hyperparameter.}
\label{tab:lambda}
\end{table*}
\subsubsection{Influence of the mixing proportion}
In this section, we perform an ablation study on four ZSL datasets. We evaluate our model with different values of the random mixing proportion $\lambda$ (Results are shown in Table \ref{tab:lambda}). Note that when $\lambda$ is sampled from a distribution, a new value is selected for each minibatch. 

We found that the best performances accross all datasets were met when $\lambda = 0.5$ or $\lambda \sim \mathcal{N}(0.5, 0.25)$, i.e. setting equal weights for the two terms of the convex combination. 
Other settings for $\lambda$, such as $\lambda = 0.2$ or $\lambda \sim Uniform(0, 1)$, deteriorates the performances ($\sim$ 1$\%$ worse for CUB, SUN and FLO and $\sim$ 4$\%$ worse for AWA2). Even poorer performances were found when setting $\lambda \sim Beta(0.3, 0.3)$ ($\sim$ 1$\%$ worse for CUB and SUN and $\sim$ 8$\%$ worse for FLO and AWA2).

For image classification \cite{mixupICLR2017}, the random mixing proportion is sampled from the Beta distribution with a small value of $\alpha$, as it assumes that the examples in the neighborhood of each data sample share the same class. Indeed, 
given a small $\alpha = 0.3$, beta distribution samples more values closer to either 0 and 1, making the mixing result closer to either one of the two examples. However, in our method, we construct \textit{ambiguous} semantic prototypes with the corresponding \textit{ambiguous} classes being completely distinct from the real ones. Therefore, sampling $\lambda$ from a Beta distribution, with $\alpha < 1$, is not a reasonable choice.

\subsubsection{Influence of the subset to learn virtual prototypes}\label{sec:influence_subset}

According to the nominal protocol of the proposed method, new ``frontier prototypes'' are learned by combining prototypes of both seen and unseen categories in the transductive settings.  The usage of seen prototypes in Eq.~\ref{eq:our_mix} allows to regularize the conditional latent space, such that further used unseen prototypes result in more discriminative features. If unseen prototypes are used, the space is better regularized in their neighborhood. It is particularly interesting if some unseen prototypes are not contained in the convex envelop of the seen classes. Let nevertheless note that we conducted experiments with negative $\lambda$ without getting noticeable improvements.
To evaluate the respective contribution of seen and unseen prototypes in our model, we compared the performances on CUB, while using different subsets of prototypes in Eq.~\ref{eq:our_mix}. For fair comparison, the results reported in Table~\ref{tab:ablation-subset} were all obtained with f-VAEGAN using unlabeled images at training time, such that the `s+u' results are the same as the transductive setting in Table ~\ref{tab:gzsl-sota} and \ref{tab:zsl-sota}.

In the ZSL setting, we obtain the same results whether we use all prototypes or unseen prototypes only, this makes sense since the test images are from the unseen classes only and there is no point in modelling the ambiguities with seen classes. It is also interesting to note that f-VAEGAN-D2 has a score of 74.2 in the transductive setting, while one can have have a score of 79.1 with the regularization learned with seen prototypes only. It shows that most of the improvement is due to the global regularization of the conditional latent space, rather than to a local one in the neighborhood of the prototypes used at test time.

In the GZSL setting, the results are better when the regularization is learned with unseen prototypes only rather than seen ones, but the usage of both is still above. Looking at the results on the seen and unseen classes specifically, one can see that the results are obviously better for the classes that are regularized with Eq.~\ref{eq:loss_mix}. The comparison to the results obtained by f-VAEGAN-D2 without the regularization in table~\ref{tab:gzsl-sota} (u=65.6 s=68.1 H=66.8) shows that the regularization is beneficial in any case.

\begin{table}[tb]
\centering
\begin{tabular}{l|c|ccc|}
    \multicolumn{1}{l}{} & \multicolumn{1}{l}{\textbf{ZSL}}  & \multicolumn{3}{c}{\textbf{GZSL}} \\
    \multicolumn{1}{l}{}  & \textbf{T1} & u & s & \textbf{H} \\ 
\hline
 s+u  & 80.6 & \textbf{74.2} & 70.5 & \textbf{72.3}  \\
 s & 79.1 & 69.4 & \textbf{70.9} & 70.1 \\
 u & \multicolumn{1}{l|}{\textbf{80.7}} & \multicolumn{1}{l}{72.6} & 70.0 & \multicolumn{1}{l|}{71.2} 
\end{tabular}
\caption{Performances on CUB according to the prototype training subset used to learn ambiguous prototypes ($s=$seen, $u$=unseen). For fair comparison, unlabeled images are used at training time in each case.}
\label{tab:ablation-subset}
\end{table}

\begin{table}[tb]
\centering
\begin{tabular}{llc|ccc}
\multicolumn{2}{l}{\multirow{3}{*}{}} & \multicolumn{1}{l|}{\textbf{ZSL}} & \multicolumn{3}{c}{\textbf{ GZSL }} \\
\multicolumn{2}{l}{} & \textbf{T1}  & u & s & \textbf{H}    \\ \hline
IN & Vanilla   & 63.2 & 52.2 & 62.7 & 56.9 \\
   &  + ours (scratch)  & 64.9 & 57.4 & 62.6 & 59.8 \\
   &  + ours (ft) & 64.3 & 54.1 & 62.7 & 58.0 \\ \hline
TR & Vanilla & 77.2 & 69.1 & 75.1 & 72.0 \\
   &  + ours (scratch)  & 79.0 & 72.3 & 75.3 & 73.7 \\
   & + ours  (ft) & 78.5 & 71.6 & 74.9 & 73.2         
\end{tabular}%
\caption{Comparison between the vanilla \texttt{TF-VAEGAN} and that augmented with our loss (Eq.~\ref{eq:loss_mix}), on CUB dataset for both inductive and transductive ZSL/GZSL settings, either when the model is learned from scratch or fine-tuned (ft)}
\label{tab:test_with_narayan}
\end{table}

\subsection{Integration to TF-VAEGAN}\label{sec:with_tfvaegan}
Our contribution is generic and can be used in other conditional generative-based ZSL architectures. Therefore, we evaluate the generalization capabilities of our proposed method, by integrating our contribution in the \texttt{TF-VAEGAN} \cite{narayan2020latent} framework.

We first learn the model end-to-end, adding the proposed regularization. Table \ref{tab:test_with_narayan} shows the comparison on CUB, between the original \texttt{TF-VAEGAN} model and the one learned  with the proposed regularization (equation~\ref{eq:loss_mix}). Our contribution improves the performance of the vanilla \texttt{TF-VAEGAN} for both ZSL and GZSL tasks, either in inductive or transductive settings, by 1.5 to 3 points. Interestingly, in GZSL, one can note that the improvement is mainly due to an increase of the scores on unseen classes, while the ones on seen classes is almost similar to the vanilla TF-VAEGAN. It thus tends to show that our approach reduces the bias towards seen classes in the generalized context.

We conducted an additional experiment consisting in fine-tuning the generator learned by \texttt{TF-VAEGAN} with our method. To prevent the generator from losing the previously learned information, which is the marginal feature distribution, the discriminators $D_1$ and $D_2$ are trained from scratch. We again observe an improvement of the performance for ZSL and GZSL, both in inductive and transductive settings. The scores are nevertheless intermediate between those obtained by the original model and those obtained previously by learning from scratch. In both cases, the GZSL experiments show that most of the score improvement is due to a better recognition of the unseen classes, while the performances on the seen classes are similar to (or slightly below) the original model.

\section{Conclusion}
We propose a novel approach to train a conditional generative-based model for zero-shot learning. The approach improves the discriminative capacity of the synthesized features by training the generator to recognize virtual \textit{ambiguous} classes. We construct the corresponding \textit{ambiguous} class prototypes as convex combinations of the \textit{real} class prototypes and then we train the generator to recognize these virtual classes. This simple procedure allows the generator to learn the transitions between categories and thus, to better distinguish them. Our approach can be integrated to any conditional generative model. Experiments on four benchmark datasets show the effectiveness of our approach across zero-shot and generalized zero-shot learning. In most cases, the improvement is due to a better recognition of unseen classes, while the score on seen classes are maintained, which means that our approach reduces the bias towards seen classes in GZSL. However, most of the time, the score on seen classes remains higher than the one on unseen classes, showing the bias still remains to some extent.

The method is limited to create ambiguous classes from a \textit{couple} of real classes by a linear interpolation. To push further our approach, one could explore non-linear interpolation for constructing ambiguous classes, or considering more than two \textit{real} classes to construct an \textit{ambiguous} one. Note that the experiment we conducted on (linear) extrapolation did not bring interesting results. Beyond this contribution to zero-shot learning, our approach can also be beneficial to other tasks that aims at relating ambiguous visual and semantic information such as multimodal entity linking and retrieval, cross-modal retrieval or classification and more generally those in which a latent space is used for learning data features. 

\textbf{Acknowledgement}: this work relied on the use of the FactoryIA cluster, financially supported by the Ile-de-France Regional Council. HLB is partially funded by CPS4EU project funded from the H2020-ECSEL-2018-IA call – Grant Agreement: 826276 and the ANR-19-CE23-0028  MEERQAT project. CH was funded by CEA for her master internship.

\textbf{Data availability} The datasets  analysed during the current study are available from the corresponding author on reasonable request. The \href{https://github.com/hanouticelina/lsa-zsl}{code},  \href{https://drive.google.com/drive/folders/1v7pXxUEuMLrP5kObjkd3MQ5SSXZlfOD8}{features} and \href{https://drive.google.com/drive/folders/13-eyljOmGwVRUzfMZIf_19HmCj1yShf1}{finetuned} features are available online.

\bibliographystyle{plain}
\bibliography{semamb_zsl}

\section{Version of the manuscript}
\begin{itemize}
   \item 07/01/2022: original manuscript
   \item 04/02/2022: fix mNRG scores; add comparison to Chou \textit{et al}. in experiments
   \item 30/05/2022: revised manuscript
\end{itemize}
\end{document}